\documentclass[sigconf]{aamas} 

\setcopyright{none}
\renewcommand\footnotetextcopyrightpermission[1]{}
\usepackage{balance} 
\usepackage{multirow}

\title{Guiding Reinforcement Learning Using Uncertainty-Aware Large Language Models}

\author{Maryam Shoaeinaeini}
\affiliation{
  \institution{University of Kentucky}
  \city{Lexington}
  \country{United States}}
\email{maryam.shoaei@uky.edu}

\author{Brent Harrison}
\affiliation{
  \institution{University of Kentucky}
  \city{Lexington}
  \country{United States}}
\email{brent.harrison@uky.edu}

\begin{abstract}
Human guidance in reinforcement learning (RL) is often impractical for large-scale applications due to high costs and time constraints. Large Language Models (LLMs) offer a promising alternative to mitigate RL sample inefficiency and potentially replace human trainers. However, applying LLMs as RL trainers is challenging due to their overconfidence and less reliable solutions in sequential tasks. We address this limitation by introducing a calibrated guidance system that uses Monte Carlo Dropout to enhance LLM advice reliability by assessing prediction variances from multiple forward passes. Additionally, we develop a novel RL policy shaping method based on dynamic model average entropy to adjust the LLM’s influence on RL policies according to guidance uncertainty. This approach ensures robust RL training by relying on reliable LLM guidance. To validate our contributions, we conduct extensive experiments in a Minigrid environment with three goals in varying environment sizes. The results showcase superior model performance compared to uncalibrated LLMs, unguided RL, and calibrated LLMs with different shaping policies. Moreover, we analyze various uncertainty estimation methods, demonstrating the effectiveness of average entropy in reflecting higher uncertainty in incorrect guidance. These findings highlight the persistent overconfidence in fine-tuned LLMs and underscore the importance of effective calibration in sequential decision-making problems.
\end{abstract}

\keywords{Reinforcement Learning, Natural Language Guidance, LLM Calibration}

\begin{document}


\pagestyle{fancy}
\fancyhead{}


\maketitle 


\section{Introduction}

Reinforcement learning (RL) excels at solving sequential decision-making problems by optimizing policies through trial-and-error learning. 
However, RL often struggles with significant sample inefficiency \cite{yu2018towards,li2023deep}, requiring vast amounts of training episodes to achieve reasonable performance. 
This is further complicated by environments with sparse rewards \cite{eschmann2021reward, shou2020reward}, where infrequent environmental feedback makes it challenging for the agent to learn effective strategies.
Recently, interactive reinforcement learning (IRL) has addressed these problems by integrating human knowledge as guidance in the training agent's loop. 
This guidance typically takes the form of either a demonstration of optimal actions \cite{hester2018deep, chen2022policy, suay2016learning}, by providing online rewards to critique or encourage the agent's action \cite{knox2009interactively, macglashan2017interactive}, or through human preferences \cite{ouyang2022training}. 
Despite the effectiveness of these approaches, it may be difficult to provide hundreds of instances of feedback or flawless demonstrations, particularly in environments that require in-depth prior knowledge\cite{harrison2017guiding}.

Lately, large language models(LLMs) have fueled notable progress in fields like medicine \cite{thirunavukarasu2023large, savage2024large} and robotics \cite{chu2023accelerating}. 
Unlike classical language models such as LSTMs, LLMs have revolutionized natural language processing (NLP) with their advanced capabilities in context learning \cite{wei2023larger, akyurek2022learning} and reasoning \cite{wei2023larger}. 
Pretraining on vast amounts of data has endowed LLMs with extensive world knowledge, extending their applications to a wide range of tasks, including text classification, sentiment analysis, high-level task planning \cite{singh2023progprompt,du2023guiding, carta2023grounding}, and decision-making \cite{bian2023chatgpt}. 

Leveraging their reasoning and task planning capabilities, LLMs have been employed to address RL's challenges with sample inefficiency and sparse rewards \cite{lin2023learning, li2024auto,chu2023accelerating}. 
Despite significant advancements, LLMs can produce problematic and inaccurate responses, often influenced by biased training data or information not grounded in their training sources \cite{huang2023look}. 
Additionally, LLMs often suffer from overconfidence in their solutions \cite{miao2021prevent} even when they generate erroneous information. 
These challenges impose a major obstacle to effectively utilizing LLMs in critical decision-making such as medicine \cite{savage2024large} or sequential decision making \cite{wen2023large}.
Therefore, it is a crucial need to build reliable LLMs through risk evaluations, better ensuring technical precision \cite{national2019national}. 
Therefore, before integrating LLMs into RL, their responses should be calibrated, and an uncertainty rate should be provided to ensure the RL agent is not misled by overconfident, inaccurate guidance. 
While extensive AI risk assessment research exists for other AI models, risk assessments for LLMs are still in the early stages \cite{huang2023look}. To the best of our knowledge, no existing IRL model utilizes calibration techniques on a LLMs' output to improve the reliability of the advice given to an IRL system. 
Moreover, there is a notable lack of reliable uncertainty estimates necessary for assessing the trustworthiness of LLM-generated guidance.

To address these issues, we propose a calibrated LLM guidance system that uses Monte Carlo Dropout(MC Dropout) to assist RL agents in sequential decision-making environments. Additionally, we introduce a dynamic entropy-based coefficient to integrate RL policy with LLM advice, enhancing the effectiveness of correct recommendations and mitigating the negative impact of erroneous ones.

\section{Related Work}
In the IRL framework, the RL agent learns through a teacher-student interaction, where a knowledgeable human provides valuable guidance or feedback, thereby accelerating the agent's training process \cite{moreira2020deep, jagodnik2017training,knox2009interactively}. 
However, there are challenges in using human advisors in complex environments: 1) Gathering sufficient human guidance is both time-consuming and expensive \cite{chu2023accelerating, li2022pre,warnell2018deep,macglashan2017interactive}; 2) Providing high-quality and flawless demonstrations is often unattainable in certain tasks due to their complexity, making it difficult for humans to determine which demonstrations will most effectively contribute to the agent's learning \cite{lin2020review, tasrin2021influencing}; and 3) Designing a hard-coded reward function is difficult in some applications \cite{arumugam2019deep, knox2012reinforcement}, as it can lead to biased behavior and suboptimal performance. To address these challenges, recent research has explored the potential of large language models (LLMs) as promising alternatives to human trainers in the RL loop.

\textbf{LLM-Enhanced RL Framework and Challenges: }With the advent of large language models (LLMs) like GPT-4\cite{achiam2023gpt} and BERT\cite{Devlin2019BERTPO}, which are trained on vast datasets and contain billions of parameters, there is now potential to overcome these RL challenges. These models can enhance sample efficiency by providing contextual guidance \cite{lin2023learning} and tackle sparse rewards by designing more effective reward functions \cite{li2024auto}.
Further, employing large language model (LLM) supervision instead of human supervision offers several advantages: it reduces the time and costs associated with human intervention, ensures consistent and high-quality guidance, and provides immediate accessibility to vast amounts of knowledge, which would be impractical to gather from human demonstrations. LLMs can serve as decision-makers, reward designers, information processors, and generators of explainability in RL \cite{cao2024survey}. They can be applied either as direct decision-makers \cite{janner2021offline, shi2023unleashing, li2022pre} by directly learning what decision an agent should take or as indirect decision-makers \cite{yao2020keep} by doing things to simplify the learning problem like generating candidate actions. 

However, this runs the risk of reducing the effectiveness of the learner if the LLM has poor performance \cite{yao2020keep}. Additionally, while LLMs are effective for real-time feedback in single-task environments, they struggle with the complexity of sequential multi-task problems \cite{chu2023accelerating}. This underscores the need for a reliable LLM guidance system that can assist RL agents without overlooking crucial actions, suggesting that LLMs are better suited as guidance systems rather than direct decision-makers or basic evaluative feedback providers.
However, even as a guidance tool, LLM-generated policies can be error-prone, especially in complex environments, making calibration and uncertainty prediction essential to improve their reliability.

\begin{figure}[t]
   \centering
   \includegraphics[width=0.35\textwidth]{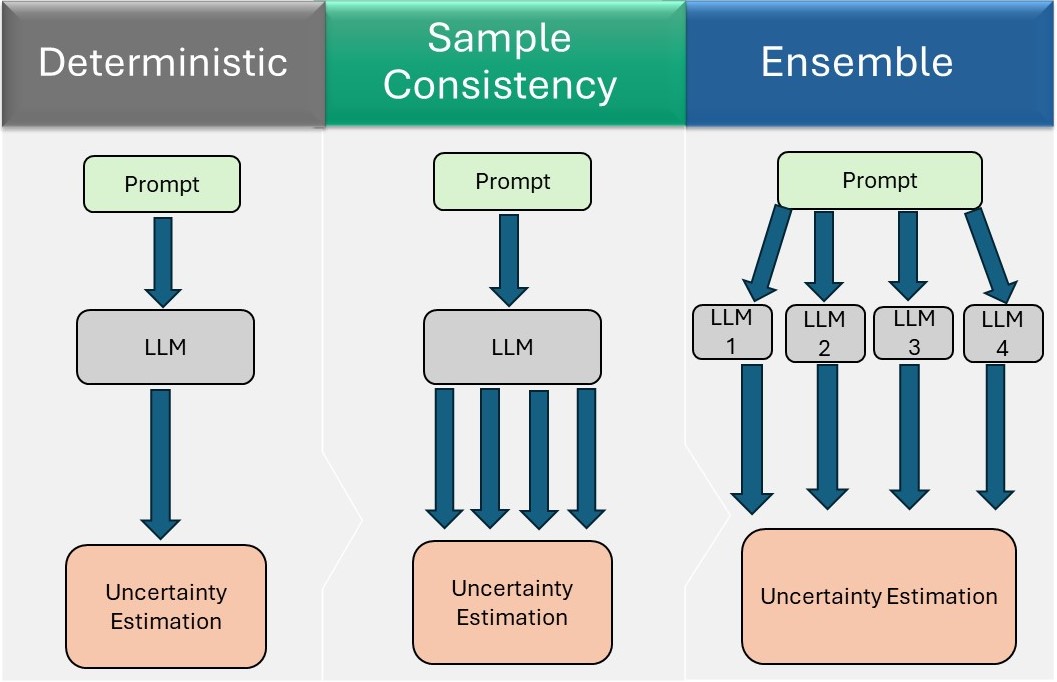}
   \caption{Illustraition of three types of uncertainty estimation methods.}
   \label{fig:UncertaintyEstimation}
\end{figure}

\textbf{LLM Calibration Techniques:}
Measuring uncertainty can be useful for identifying incorrect responses in various NLP tasks \cite{huang2023look}. Generally, machine learning models encounter two primary forms of uncertainty in their predictions: aleatoric uncertainty and epistemic uncertainty \cite{kendall2017uncertainties}. Aleatoric uncertainty is related to observation errors, such as sensor noise, while epistemic uncertainty arises from limited knowledge about the model's parameters, often because of insufficient training data. 
Although LLMs demonstrate remarkable capabilities and rapid advancements, they frequently produce incorrect information unexpectedly, including hallucinations \cite{ji2023survey}, disinformation \cite{tamkin2021understanding}, or bias \cite{abid2021persistent}.
Due to fine-tuning the LLM on the specific task, we primarily focus on epistemic uncertainty in this paper. As illustrated in Figure \ref{fig:UncertaintyEstimation}, uncertainty estimation methods can be categorized into three types: deterministic \cite{oberdiek2018classification}, sample consistency \cite{barber1998ensemble}, and ensemble approaches \cite{lakshminarayanan2017simple,xiao2021hallucination}. 

Since we are focused on determining the uncertainty of a single LLM, we only use deterministic and sample consistency methods in this work. 
Deterministic methods measure uncertainty using a single forward pass of a model. Deterministic methods like logit-based methods (log-probabilities) \cite{guo2017calibration, jiang2021can} or entropy-based methods \cite{huang2023look} are mainly applied in classification tasks.
Among the deterministic methods, average entropy outperforms other uncertainty estimation methods based on the result obtained in the question-answering task by \cite{huang2023look}. 
Sample consistency methods utilize randomness in a model's parameters (such as Bayesian methods) or data (like test-time data augmentation) to generate a collection of non-deterministic predictions and estimate uncertainty based on variation in predictions. 

Specifically, our work builds on \cite{felicioni2024importance}, which shows that uncertainty estimation outperforms the greedy approach in a basic contextual bandit problem. Our study explores the effectiveness of various epistemic uncertainty methods, deterministic and sample consistency calibrations, compared to the uncalibrated greedy approach in LLM-based RL as an indirect decision-maker. We specifically examine these methods in the context of scaling to sequential multi-task environments and introduce a novel method to integrate LLM guidance with agent policy based on the uncertainty of each guidance.

\section{Background}

\subsection{Reinforcement Learning}
Reinforcement Learning (RL) is a learning algorithm in which an agent interacts with a learning environment to optimize its policy using environmental feedback in a trial-and-error process. 
In RL, each trajectory consists of numerous steps where the agent takes action \( a \) based on the observed state \( s \) and receives the reward \( r \) from the environment. The agent aims to maximize the cumulative rewards over each trajectory by optimizing its policy $\pi$. The optimization problem is characterized by the Markov Decision Process (MDP) concept, expressed as a quintuple of \( < S , A, T, R , \gamma > \). 
In this structure, \( S \) and \( A \) denote set of all potential states and actions, respectively.
\( T \) specifies the agent transitioning function from one state to another \( T : S \times A \times S \rightarrow [0, 1]\). \( R \) is a reward function, \( R: S \times A \rightarrow \) $\mathbb{R}$. 
Lastly, the discount factor, denoted by $\gamma$, determines the importance of future rewards relative to immediate rewards. 

In this work, we implement our techniques on top of a Proximal Policy Optimization (PPO) algorithm \cite{schulman2017proximal}. 
PPO stands out among on-policy methods for its ability to provide more reliable action probabilities through its unique clipping technique. This clipping mechanism helps to maintain stability in the training process, preventing the policy from diverging or oscillating significantly. 



\subsection{Policy Shaping}
In this work, we also utilize concepts found in policy shaping. 
Policy shaping in RL is a technique where external guidance, typically provided by a human expert or a trained AI system, is integrated into the learning process to influence or shape the agent's policy~\cite{griffith2013policy,cederborg2015policy}.
This is typically done by maintaining two action distributions, one that represents the agent's policy based on its own experience and another that is created based on user feedback. 
These two distributions are then combined during action exploration to create a combined policy that the agent uses to guide its exploration. 
Typically, this is done through a weighted pointwise multiplication. 
These techniques, which aim to influence an agent's actions, tend to be more effective than those that alter the agent's rewards and value functions \cite{yu2018learning}, such as reward shaping.

\subsection{LLM's Customization}
To effectively customize an LLM for specific tasks, there are four main methods: prompt engineering, Retrieval-Augmented Generation (RAG), fine-tuning, and pretraining \cite{rag}. In this work, we focus on prompt engineering and fine-tuning. Prompt engineering involves crafting targeted prompts to guide the LLM's responses effectively, while fine-tuning adjusts the model's parameters based on task-specific data to enhance performance. Unlike RAG, which combines external data, or pretraining, which requires extensive computational resources, prompt engineering and fine-tuning offer practical and efficient methods to improve LLM reliability and relevance for specialized tasks.
\subsection{Uncertainty Evaluation}
Uncertainty metrics are evaluated based on two key aspects: Discrimination and Calibration \cite{savage2024large}. Discrimination assesses an uncertainty measure's ability to distinguish between correct and incorrect answers, reflecting how effectively the metric identifies the accuracy of the LLM's responses. 
Calibration checks if the predicted probability of accuracy from the uncertainty metric matches the actual observed probability. 
In this study, since the agent learns through reinforcement learning (RL) and lacks a dataset, actual observed probability is not available, making calibration impossible to evaluate. 
Additionally, uncertainty does not always correspond with inaccuracy; for example, a low uncertainty level does not guarantee the reliability of an LLM’s response \cite{huang2023look}. 
An LLM can be highly confident even when it provides incorrect information. 
Therefore, to assess the effectiveness of our uncertainty measure in terms of discrimination, it is important to evaluate whether the uncertainty rate surpasses a threshold when the predicted class deviates from the actual class, which may reveal if the model is overconfident.

Uncertainty estimation methods are evaluated using two distinct metrics: Expected Calibration Error (ECE) and Brier Score (BS).
According to equation \ref{ece}, ECE measures the calibration of a model by quantifying the difference between predicted confidence and actual accuracy. 
In this equation, \( M \) is the number of bins, \( B_m \) is the set of indices of samples whose predicted confidence scores fall into the \( m \)-th bin, \( n \) is the total number of samples, \( \mathrm{acc}(B_m) \) is the accuracy of the samples in bin \( B_m \), and \( \mathrm{conf}(B_m) \) is the average confidence of the samples in bin \( B_m \).
This aggregated metric provides a single value representing how well the predicted probabilities align with true outcomes.

On the other hand, BS as defined in equation \ref{bs}, measures the accuracy of probabilistic predictions by calculating the mean squared difference between predicted probabilities and actual outcomes.
Here, \( N \) represents the total number of predictions, \( f_i \) denotes the predicted probability for the \( i \)-th prediction, and \( o_i \) indicates the actual outcome for the \( i \)-th prediction, where \( o_i \) is 1 if the event occurred and 0 otherwise.
In our classification problem, \( o_i \) if the LLM prediction matches the oracle prediction, and \( f_i \) is either 1-mean entropy or max(probability), applicable to both deterministic and sample consistency experiments.
For both ECE and BS, lower values indicate better calibration and prediction accuracy.

\begin{eqnarray}
    \mathrm{ECE} = \sum_{m=1}^M \frac{|B_m|}{n} \left| \mathrm{acc}(B_m) - \mathrm{conf}(B_m) \right|
\label{ece}
\end{eqnarray}

\begin{eqnarray}
    \mathrm{BS} = \frac{1}{N} \sum_{i=1}^N (f_i - o_i)^2
\label{bs}
\end{eqnarray}

\section{Calibrated LLM Trainer Structure}
This section describes the model architecture, starting with the calibration framework for LLM guidance and prediction uncertainty. It then explains how this calibrated guidance integrates with the agent's policy and how the agent learns using the PPO reinforcement learning method.

\subsection{Calibration System Framework}
To address the issue of miscalibration in LLMs, we present a calibration system based on MC Dropout.
This system enhances the reliability of LLMs by estimating and calibrating their uncertainty. 
By incorporating MC Dropout, we perform multiple stochastic forward passes during inference, generating several action probability distributions rather than a single action probability distribution. 
These distributions provide valuable insights into the model's confidence and uncertainty in its predictions. 

Our Calibration system operates as follows:
1)	Input Data Preparation: The input data is preprocessed and tokenized using the fine-tuned tokenizer of the LLM. The resulting tokens are then fed into the model.
2)	MC Dropout: During inference, we activate dropout layers of the fine-tuned LLM. This technique prevents overfitting, leading to induced variability in the model’s predictions.
3)	Stochastic Forward Passes: Multiple forward passes are conducted, each with a different set of dropped-out neurons. Each pass results in an action probability distribution.
4)	Aggregating Action Probabilities: For each action, average the probabilities across all forward passes to obtain a calibrated probability distribution. This results in a single, averaged action probability distribution.
5)	Calculating Entropy: Compute the entropy of the averaged action probability distribution to quantify the model's uncertainty.

\begin{figure}[h]
    \centering
    \includegraphics[width=0.45\textwidth]{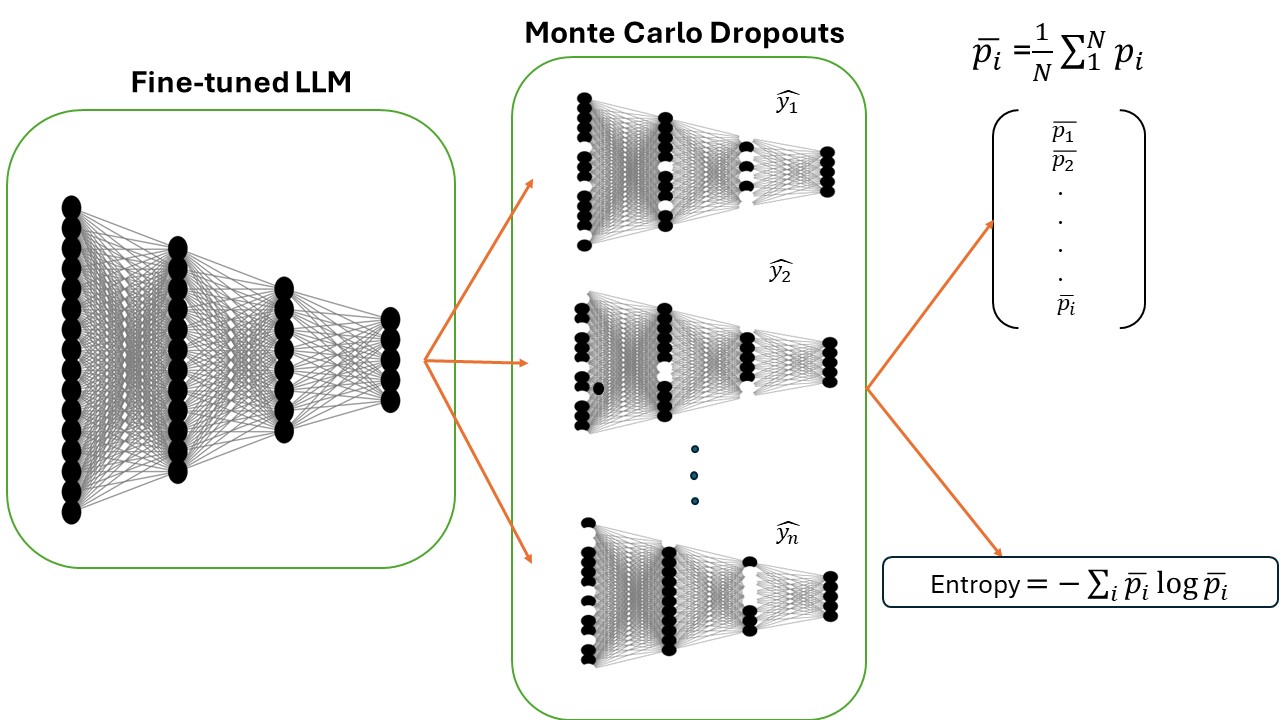}
    \caption{The calibration system architecture using MC Dropout in the fine-tuned LLM.}
    \label{fig:MonteCarlo_Framework}
     \Description{.}
\end{figure}

As shown in Figure \ref{fig:MonteCarlo_Framework}, neurons in the fine-tuned LLM network are randomly dropped based on the dropout rate. This process effectively creates several sub-networks, each with a different set of dropped neurons. The same prompt is then passed through each of these sub-networks, resulting in multiple action probability distributions. Subsequently, each generated piece of advice has a specific uncertainty rate determined by the entropy of the averaged probability distribution from the multiple forward passes. Lower entropy indicates higher confidence, while higher entropy signifies greater uncertainty. This allows users to gauge the reliability of the given advice and evaluate the model's performance over time.
\subsection{LLM-Enhanced RL Architecture}
\begin{figure}[h]
    \centering
    \includegraphics[width=0.9\linewidth]{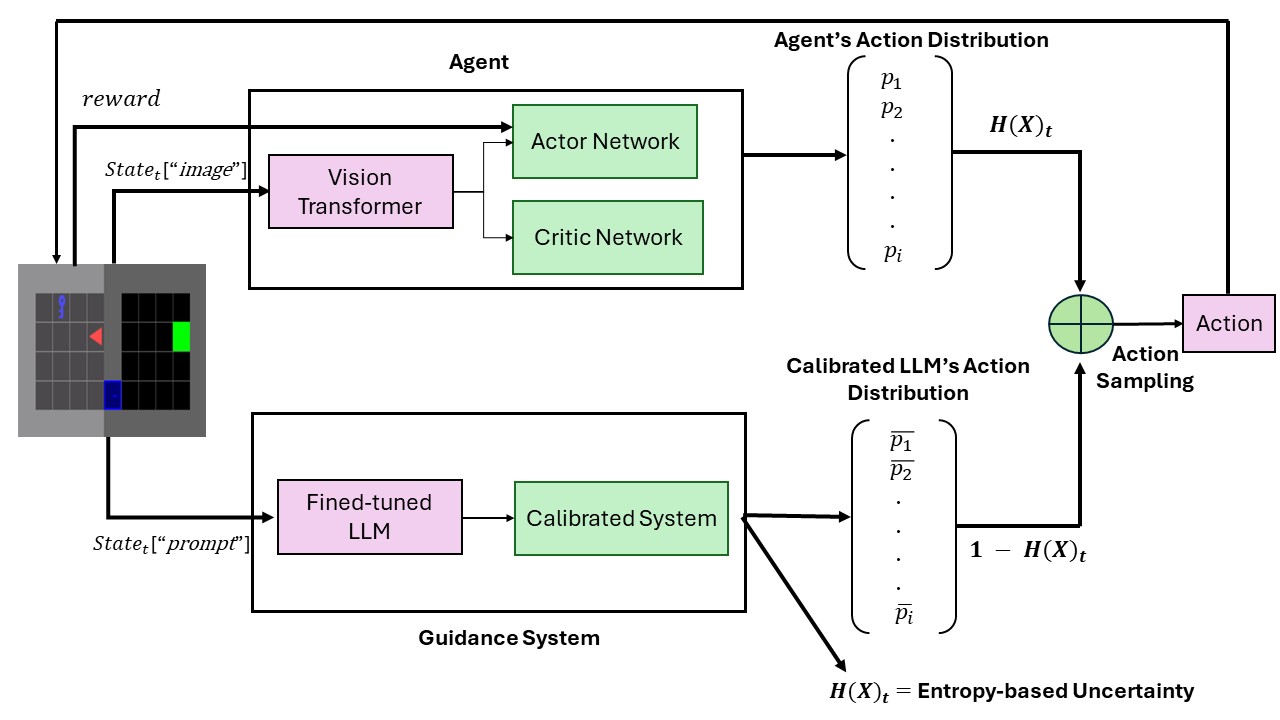}
    \caption{The structure of calibrated LLM-based RL system }
    \label{fig:LLM_enhanced_RL}
      \Description{.}
\end{figure}

Our guidance model is implemented by fine-tuning a pretrained LLM for a downstream sequential multi-task environment. The LLM serves as an indirect decision-making system, enhancing the sample efficiency of online reinforcement learning (RL) by leveraging its robust sequence modeling capabilities and common sense knowledge. Leveraging transformers in the architecture, the agent processes the state's image through a vision transformer and the state's prompt using a fine-tuned LLM, as illustrated in Figure \ref{fig:LLM_enhanced_RL}. The embedding layer retrieved from the vision transformer is fed into the actor and critic networks to generate the agent's action probability distribution and expected return, respectively. Meanwhile, a dynamic prompt showing the environment context and the agent's mission at time step \( t \) is fed into the guidance system. The pretrained LLM is fine-tuned through an oracle to provide valuable guidance more often. However, due to miscalibration and over-parameterization of the LLM, guidance can become distracted and inaccurate, especially in long-horizon and multi-task environments.
To address this issue, the guidance system is calibrated for each piece of advice through a calibration system. Thus, the generated action probability distribution from the guidance system is calibrated. The entropy of the calibrated advice, denoted by \( H(X)_t \) indicates the uncertainty of the guidance system for that advice. Instead of using a constant coefficient to integrate the agent's action distribution and calibrated advice, we use the dynamic uncertainty of the guidance system as shown in equation \ref{policy}. This approach provides the agent with more informed and reliable advice, allowing it to outperform the guidance system based on the learned policy in cases of uncertain advice over time.
\begin{eqnarray}
    P_a(t) = (1 - H(X)_t) \times P_{\text{LLM}}(t) + H(X)_t \times P_{\text{Agent}}(t)
\label{policy}
\end{eqnarray}
Here, \( P_{\text{LLM}}(t) \) denotes the probability distribution of actions predicted by the calibrated LLM, \( P_{\text{Agent}}(t) \) represents the probability distribution of actions based on the agent's policy, and \( P_t(a) \) indicates the combined action probability distribution at time \( t \).

\section{Experimental Setup}
To assess the calibration system's effectiveness, we compare calibrated LLM-based RL using MC Dropout with an uncalibrated deterministic approach, focusing on how calibration improves reliability and accuracy when the same prompt is repeatedly passed multiple times to a single LLM. Additionally, we demonstrate the benefits of uncertainty-aware policy shaping by comparing it to a linear decay coefficient method, where the influence of the LLM's action probability distribution decreases from 1 to 0 over the course of episodes. These evaluations are conducted through several experiments in Minigrid's Unlock Pickup environment, using the same seed for each run. Maintaining a fixed seed ensures that each experiment starts with the same initial conditions, allowing for a controlled comparison of the calibration and policy shaping methods. Each model is run for 3040 episodes with the same reward settings and the PPO algorithm for online RL. The model is evaluated in two different environment sizes: 4x4 and 3x3, to assess their effectiveness across varying scales comprehensively. In all experiments, we report the average episode reward after smoothing it using a moving window of 250 episodes. 
\subsection{Fine-Tuning: }
Before conducting the experiments, the LLM is fine-tuned using an oracle system that assumes a single optimal action at each step. To introduce diversity into the fine-tuning dataset, the agent is given either a random action or the oracle's action during training. However, the dataset is specifically curated to include only states where the oracle's action was selected, ensuring robust fine-tuning in the complex sequential multi-task environment. In this study, we fine-tune the BERT language model using a dataset of 21,500 states for a 4x8 grid environment, achieving 90\% accuracy in the final evaluation. For a smaller 3x6 grid environment, BERT is fine-tuned with a dataset of 15,000 states, resulting in 93\% accuracy in the final evaluation.




\subsection{Model Parameters}
In the calibration system's structure, the dropout probability is set to 0.1, aligning with the inference LLM dropout rate used in \cite{felicioni2024importance} and matching the rate at which BERT was pre-trained. To generate robust fine-tuning LLM, the dropout is utilized only in the inference phase. For the actor and critic networks in the RL agent, the embedding layer size is fixed at 1024. The PPO algorithm is updated using the Adam optimizer, with a learning rate of $10^{-4}$, a batch size of 15, and 4 epochs.

\subsection{Environment and Reward Structure}
The Minigrid unlock pickup environment is a gridworld in which an agent must pick up a key to unlock a door to leave the environment. 
Using MDP terminology, the state, \( S_t \) is composed of an image of the state and a natural language prompt that includes the agent's current state and goal information. The action space\( A_t \) is defined as: 0 (going left), 1 (going right), 2 (going straight), 3 (picking up the key), and 5 (opening the door). As illustrated in \ref{fig:prompt_example}, the agent must first pick up the key, then open the door, and finally proceed to the green goal in that specific order. 

Transitions \( T_t \) between states \( S_t \) and \( S_t+1 \) are determined by the action \( a_t \) taken by the agent at time \(t\). The environmental reward \( r_t \) is given only upon completing the final mission, calculated as \(\ 1 - \frac{\text{step\_count}}{\text{max\_steps}}\) 
, with values ranging from 0 to 1. This sparse reward function can cause sample inefficiency by promoting the exploration of less relevant states. We assign constant rewards for completing each completed task to mitigate the challenges of multi-tasking and sparse rewards. In the updated reward function setting, the agent receives a reward of 0.5 for completing the first mission and another 0.5 for accomplishing the second mission. For the third mission, the agent receives an additional 0.2 on top of the environmental reward to balance the rewards between missions and encourage sequential task completion. Furthermore, due to the positive rewards, the agent may over-prioritize certain actions, such as picking up the key (action 3) or opening the door (action 5), even when these actions are performed in incorrect states. To mitigate this issue, we introduce a small negative reward of -0.02 for performing these actions in inappropriate states. This adjustment helps the agent learn to balance actions effectively over time.  

\subsection{Prompt Engineering}
\begin{figure}[h]
    \centering
    \includegraphics[width=0.3\linewidth]{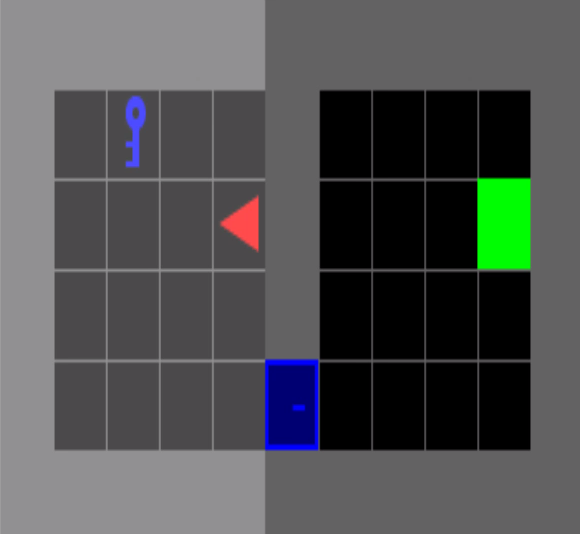}
    \caption{An instance of agent's state}
    \label{fig:prompt_example}
     \Description{.}
\end{figure}

The state's prompt is designed to provide the environmental context and the agent's mission at each time step. This prompt dynamically changes based on the position of the agent and the specific mission objectives.
Below, for example, is the prompt associated with the state shown in Figure \ref{fig:prompt_example}.

\textit{prompt: The red agent is in a 4x4 grid environment surrounded by walls. Each grid cell is identified by coordinates (i, j), where i denotes the column and j denotes the row. The agent can turn left (action 0), turn right (action 1), move forward (action 2), pick up key (action 3), and open door (action 4). The agent can face right (0), down (1), left (2), or up (3). The agent cannot pass through walls. It can open the door if it has the key and is facing the closed door, and it can pick up the key when facing it. The agent needs to find the shortest route to key or door and then pickup the key or open the door. Consider the direction as the way the agent is facing, not the way we are seeing the agent, to avoid mixing right and left. In this state, the agent is at position (4, 2), the agent direction is < and agent's direction number is 2, and the forward object is empty cell, and the key position is (2, 1), the key is not being carried by the agent, the door is at position (4, 3), the goal is at position (5, 1), the door is False open, and the mission is pick up key. What is the optimal action for the agent to take in this state to accomplish the mission?just say the optimal action number} 

\section{Results}
As discussed, we compare the proposed model's performance against the following baselines: RL without LLM guidance (unguided RL), uncalibrated LLM-enhanced RL, and a model using linear policy shaping. 
We also perform these experiments in a 4x4 environment and a 3x3 environment to investigate how these approaches scale with the size of the state space. Finally, we analyze the discrimination and calibration of uncertainty estimation methods based on the experimental results.

\begin{figure}[h]
    \centering
    \includegraphics[width=0.8\linewidth]{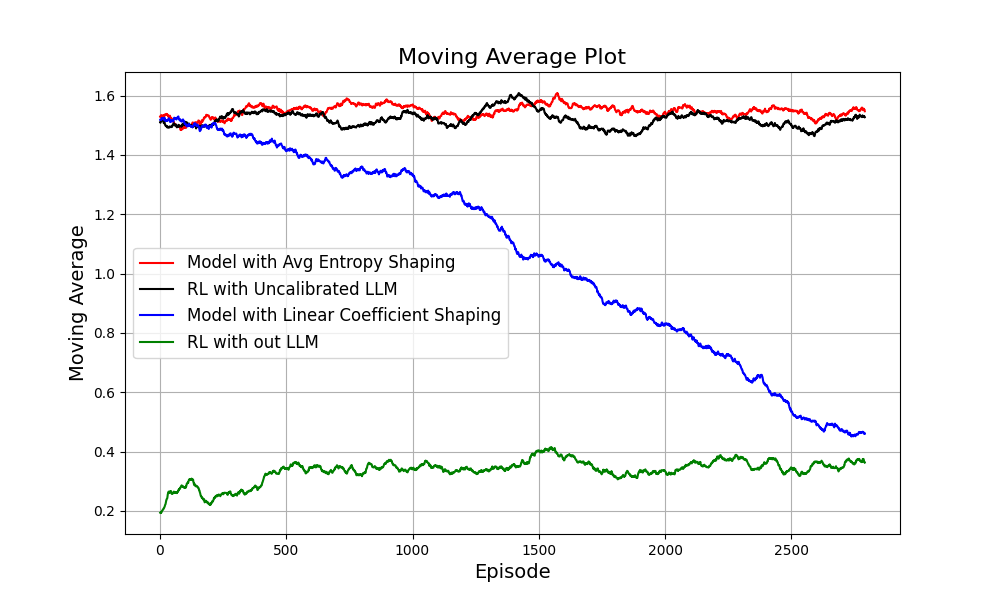}
    \caption{Comparison of four models—uncalibrated LLM guided, unguided RL, linear policy shaping, and our uncertainty-aware policy shaping model.}
    \label{fig:noLLM_Comparison__2_.png}
     \Description{.}
\end{figure}

\textbf{Comparison to RL without Guidance: }To assess the benefits of incorporating a guidance system, we compare our model with an RL agent operating without guidance, serving as our baseline. The baseline relies solely on a traditional online reinforcement learning algorithm. Figure \ref{fig:noLLM_Comparison__2_.png} highlights the comparative performance of our model, demonstrating its superior sample efficiency. In the experiments, the average reward for the unguided RL agent levels off at around 0.4, indicating limited improvement with additional training. In contrast, the calibrated LLM-enhanced RL agent achieves a plateau around 1.6, showcasing significantly better performance and higher sample efficiency. This comparison underscores the effectiveness of the guidance system in enhancing the learning process and achieving better results.

\textbf{Comparison to Uncalibrated LLM-Enhanced RL}
To emphasize the robustness achieved through calibrating the LLM guidance, we compare the results from both the uncalibrated LLM and the calibrated guidance system. Figure \ref{fig:noLLM_Comparison__2_.png} illustrates the performance of these models, with the red line representing the calibrated LLM-enhanced RL and the black line representing the uncalibrated LLM-enhanced RL. The calibrated LLM-enhanced RL model outperforms the uncalibrated counterpart, as evidenced by a higher area under the curve. This enhanced performance is further detailed in Table \ref{auc}, where the superior results of the calibrated guidance system are quantified. These findings demonstrate that calibration significantly improves the robustness and effectiveness of the LLM guidance in reinforcement learning tasks.

\begin{table}[t]
  \caption{Area Under Curve (AUC) metric for all experiments.}
  \label{auc}
  \begin{tabular}{ll}\toprule
    \textit{Method} & \textit{AUC} \\ \midrule
    Our Model & \textit{4318.65} \\
    Unguided RL & 938.24 \\
    Uncalibrated LLM-Enhanced RL & 4240.22 \\
    Calibrated LLM-Enhanced RL by Decay Coefficient & 2977.79 \\
    Our Model in 3x3 Environment & 4194.17 \\
    \bottomrule
  \end{tabular}
\end{table}

\textbf{Comparing the Uncertainty-Aware Policy Shaping Method with Linear Decay Coefficient: }To effectively integrate a guidance system with reinforcement learning, we employ a dynamic entropy-based coefficient for policy shaping, as the uncertainty of LLM advice varies at each state. This approach is compared against a baseline method using a fixed linear decaying coefficient. In the baseline method, the LLM's coefficient starts at 1 and linearly decreases to 0 by the final episode. As shown in Figure \ref{fig:noLLM_Comparison__2_.png}, the guidance system, when combined with a fixed linear decaying factor, fails to perform efficiently in sequential multi-tasks environment. The fixed approach lacks the adaptability needed to account for varying levels of uncertainty in the LLM's guidance. In contrast, our proposed model, which uses a dynamic entropy-based policy shaping method, maintains an upward performance trend. This method effectively leverages the guidance system by adjusting the policy shaping coefficient based on the uncertainty at each state, leading to more efficient and robust learning outcomes.

\begin{figure}[h]
    \centering
    \includegraphics[width=0.8\linewidth]{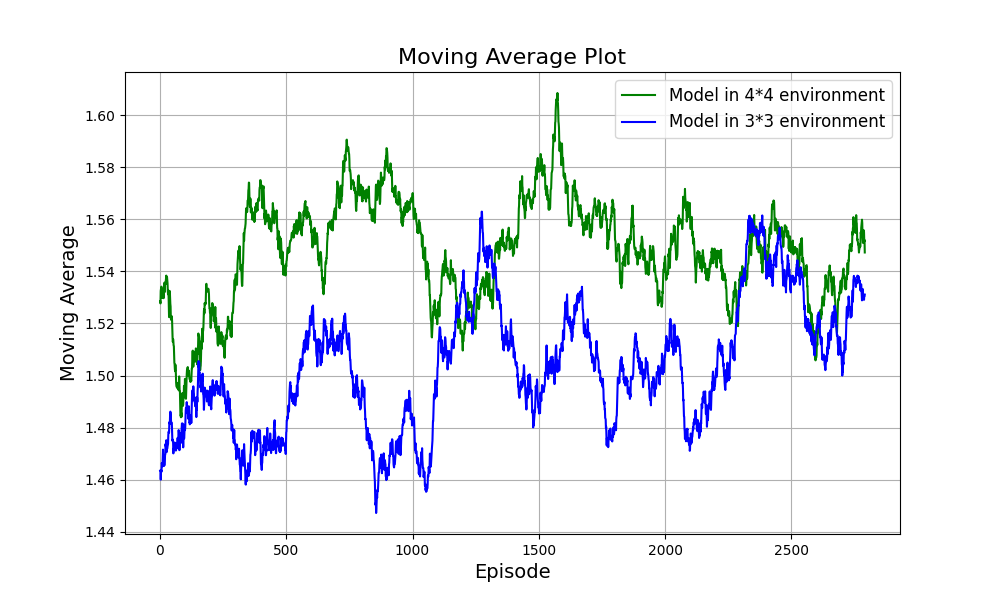}
    \caption{Comparison of calibrated model performance in different environment sizes.}
    \label{fig:env_sizes}
     \Description{.}
\end{figure}

\textbf{Comparison to Smaller Environment Size: }To assess the performance of calibrated BERT in smaller and simpler environments, we conducted experiments in a 3x3 grid environment with identical tasks. Surprisingly, as shown in Figure \ref{fig:env_sizes}, the model performed better in the 4x4 environment than in the 3x3 environment. This unexpected result is due to the higher overconfidence of the LLM in the more accessible 3x3 environment compared to the more challenging 4x4 environment. Additionally, the 4x4 environment offers a greater variety of states, allowing the agent to learn more effectively. This suggests that the model's ability to calibrate and assess uncertainty may be influenced by the complexity of the environment, affecting its predictive accuracy. Future work can explore how overconfidence in LLMs varies with different environment scales to enhance performance. 

\textbf{Analysis of Discrimination of Uncertainty Metrics: } As previously noted, uncertainty estimation does not always align with prediction accuracy in LLMs. To address this, we assess the discrimination capability by analyzing instances of model overconfidence in incorrect predictions. Specifically, we compare how often uncertainty estimates exceed 50\% (using a random estimation baseline) when the predicted class differs from the oracle's class, across various uncertainty estimation methods. As shown in Table \ref{tab2}, average entropy demonstrates the highest discrimination, achieving 80\% in the sample consistency method compared to other scenarios.

\begin{figure}[h]
    \centering
    \includegraphics[width=0.60\linewidth]{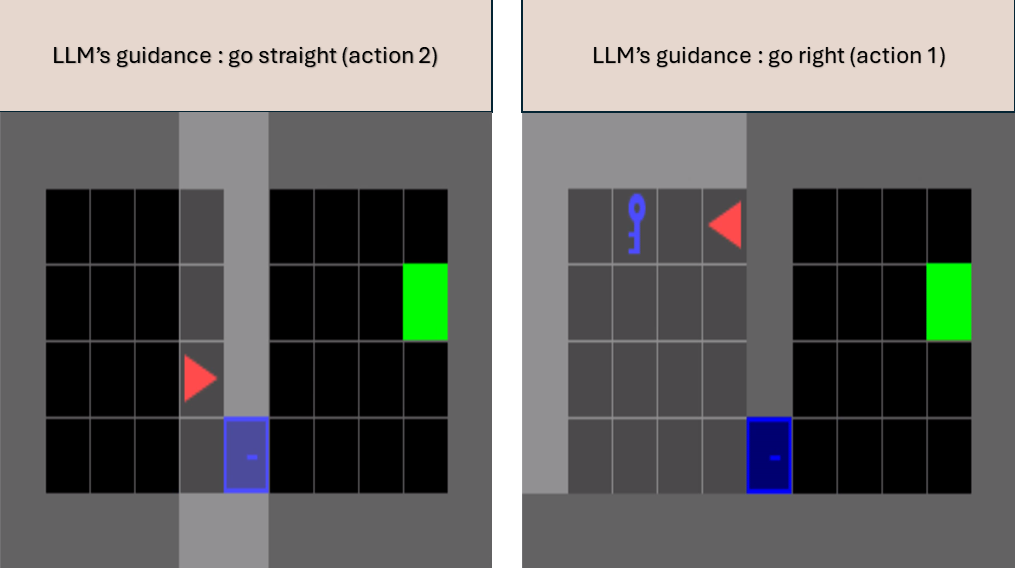}
    \caption{Instances of incorrect advice: the calibrated LLM (right) and the uncalibrated LLM (left).}
    \label{fig:Wrong_Guidance}
     \Description{.}
\end{figure}

Additionally, average entropy shows greater discrimination in the deterministic method than the 1 - maximum probability method. The average entropy uncertainty in a calibrated LLM is more robust in cases of incorrect guidance. For instance, in the right scenario depicted in Figure \ref{fig:Wrong_Guidance}, the calibrated LLM provides incorrect advice to take action 1 (going right), which is incorrect as the agent faces to the wall. Here, average entropy is 67\%, surpassing the 50\% threshold, whereas the maximum probability method shows only 38\% uncertainty. In contrast, in the left instance shown in Figure \ref{fig:Wrong_Guidance}, the uncalibrated LLM advises the incorrect action 2, which is incorrect as the agent cannot pass through the wall, yet the average entropy is only 23\%. This demonstrates the superior reliability of the calibrated LLM in signaling uncertainty when giving incorrect advice.

\textbf{Analysis of Calibration Methods: } The results of calibration metrics for deterministic and sample consistency methods, using average entropy and 1 - maximum probability, are reported in Table \ref{tab2}. The reliability of average entropy as an uncertainty metric for calibrated BERT models is evidenced by its lower ECE and Brier Score across both environment sizes, compared to 1 - maximum probability.
Despite the small differences in ECE and Brier Score for average entropy between the deterministic method (uncalibrated LLM) and sample consistency (calibrated LLM), their significant impact is evident when comparing their performance in guiding the RL agent, as shown in Figure \ref{fig:noLLM_Comparison__2_.png}.


\begin{table}[t]
\caption{Evaluation of different uncertainty estimation methods using ECE and BS}
\resizebox{\columnwidth}{!}{%
\begin{tabular}{lllc} \toprule
\textit{Methods}  & \textit{ECE} & \textit{BS}  & \textit{Discrimination}\\ \midrule
Deterministic 4*4 by Mean Entropy & 0.16 & 0.21 & 0.76 \\ 
                        Deterministic 4*4 by Max Probability & 0.27 & 0.27 & 0.74 \\ 
Sample Consistency 4*4 by Mean Entropy & \textbf{0.15} & \textbf{0.20} & \textbf{0.8}\\ 
Sample Consistency 4*4 Max Probability & 0.26 & 0.26 & 0.74 \\
Sample Consistency 3*3 by Mean Entropy & 0.14 & 0.19 & 0.75\\ 
Sample Consistency 3*3 by Max Probability & 0.19 & 0.22& 0.75 \\ \bottomrule
\end{tabular}%
}
\label{tab2}
\end{table}

\section{Discussion}
\textbf{Can LLMs replace human trainers in guiding RL agents?}  To gauge the effectiveness of LLMs as trainers and enhance the reliability of their guidance through MC Dropout calibration, we fine-tuned a BERT model. We conducted extensive experiments in a Minigrid environment with three sequential tasks. Our results show that fine-tuned LLMs significantly boost RL agent performance, achieving an average reward of 1.6 compared to 0.4 for unguided RL agents, with a difference of 3,380.41 in the area under the curve. The calibrated guidance system also demonstrated superior performance over the uncalibrated version, resulting in more robust training and higher average rewards. Interestingly, using the model in smaller, simpler environment led to increased overconfidence and reduced performance, indicating that calibration and uncertainty assessment may be affected by the complexity of the environment.

\textbf{How can LLMs be integrated to shape RL policy for ensuring robust guidance?}  Incorporating uncertainty into policy shaping significantly enhances the efficacy and robustness of RL training. Unlike linear policy shaping, which is challenging to optimize for different problems, using average entropy provides an efficient and automatic balance between the LLM and the RL agent. Our novel uncertainty-aware policy shaping method outperforms traditional linear decay weight methods, achieving a 45\% increase in the area under the curve for training rewards.
Additionally, an analysis of estimation metrics showcased the superior discrimination accuracy of average entropy in the sample consistency method, consistently exceeding 50\% in most instances of incorrect guidance. Conversely, average entropy uncertainty showed lower discrimination accuracy in deterministic calibration, highlighting the importance of the multiple forward pass method for effective calibration.

\section{Conclusion}
In this paper, we propose an uncertainty-aware LLM-enhanced RL framework that simultaneously reduces LLM overconfidence and improves RL sample efficiency. By applying MC Dropout during the inference stage of a fine-tuned BERT model, the calibrated BERT demonstrated superior performance compared to RL with uncalibrated LLMs and RL without LLMs in a sequential multi-task Minigrid environment. Additionally, our novel uncertainty-aware policy shaping method contributed to maintaining an upward training reward trend, in contrast to the downward trend observed with traditional policy shaping methods like linear decaying coefficients. Moreover, Discrimination analysis proved the efficiency of sample consistency calibration methods compared to deterministic ones. Notably, among uncertainty metrics, the model's average entropy demonstrated superior performance in reflecting higher uncertainty (over 50\%) in incorrect guidance, revealing its potential in mitigating LLM overconfidence. These promising findings pave the way for further advancements in LLM-in-the-loop RL systems centered on LLM uncertainty.




\bibliographystyle{ACM-Reference-Format} 
\bibliography{sample}


\begin{thebibliography}{56}


\ifx \showCODEN    \undefined \def \showCODEN     #1{\unskip}     \fi
\ifx \showDOI      \undefined \def \showDOI       #1{#1}\fi
\ifx \showISBNx    \undefined \def \showISBNx     #1{\unskip}     \fi
\ifx \showISBNxiii \undefined \def \showISBNxiii  #1{\unskip}     \fi
\ifx \showISSN     \undefined \def \showISSN      #1{\unskip}     \fi
\ifx \showLCCN     \undefined \def \showLCCN      #1{\unskip}     \fi
\ifx \shownote     \undefined \def \shownote      #1{#1}          \fi
\ifx \showarticletitle \undefined \def \showarticletitle #1{#1}   \fi
\ifx \showURL      \undefined \def \showURL       {\relax}        \fi
\providecommand\bibfield[2]{#2}
\providecommand\bibinfo[2]{#2}
\providecommand\natexlab[1]{#1}
\providecommand\showeprint[2][]{arXiv:#2}

\bibitem[\protect\citeauthoryear{Abid, Farooqi, and Zou}{Abid et~al\mbox{.}}{2021}]%
        {abid2021persistent}
\bibfield{author}{\bibinfo{person}{Abubakar Abid}, \bibinfo{person}{Maheen Farooqi}, {and} \bibinfo{person}{James Zou}.} \bibinfo{year}{2021}\natexlab{}.
\newblock \showarticletitle{Persistent anti-muslim bias in large language models}. In \bibinfo{booktitle}{\emph{Proceedings of the 2021 AAAI/ACM Conference on AI, Ethics, and Society}}. \bibinfo{pages}{298--306}.
\newblock


\bibitem[\protect\citeauthoryear{Achiam, Adler, Agarwal, Ahmad, Akkaya, Aleman, Almeida, Altenschmidt, Altman, Anadkat, et~al\mbox{.}}{Achiam et~al\mbox{.}}{2023}]%
        {achiam2023gpt}
\bibfield{author}{\bibinfo{person}{Josh Achiam}, \bibinfo{person}{Steven Adler}, \bibinfo{person}{Sandhini Agarwal}, \bibinfo{person}{Lama Ahmad}, \bibinfo{person}{Ilge Akkaya}, \bibinfo{person}{Florencia~Leoni Aleman}, \bibinfo{person}{Diogo Almeida}, \bibinfo{person}{Janko Altenschmidt}, \bibinfo{person}{Sam Altman}, \bibinfo{person}{Shyamal Anadkat}, {et~al\mbox{.}}} \bibinfo{year}{2023}\natexlab{}.
\newblock \showarticletitle{Gpt-4 technical report}.
\newblock \bibinfo{journal}{\emph{arXiv preprint arXiv:2303.08774}} (\bibinfo{year}{2023}).
\newblock


\bibitem[\protect\citeauthoryear{Aky{\"u}rek, Schuurmans, Andreas, Ma, and Zhou}{Aky{\"u}rek et~al\mbox{.}}{2022}]%
        {akyurek2022learning}
\bibfield{author}{\bibinfo{person}{Ekin Aky{\"u}rek}, \bibinfo{person}{Dale Schuurmans}, \bibinfo{person}{Jacob Andreas}, \bibinfo{person}{Tengyu Ma}, {and} \bibinfo{person}{Denny Zhou}.} \bibinfo{year}{2022}\natexlab{}.
\newblock \showarticletitle{What learning algorithm is in-context learning? investigations with linear models}.
\newblock \bibinfo{journal}{\emph{arXiv preprint arXiv:2211.15661}} (\bibinfo{year}{2022}).
\newblock


\bibitem[\protect\citeauthoryear{Arumugam, Lee, Saskin, and Littman}{Arumugam et~al\mbox{.}}{2019}]%
        {arumugam2019deep}
\bibfield{author}{\bibinfo{person}{Dilip Arumugam}, \bibinfo{person}{Jun~Ki Lee}, \bibinfo{person}{Sophie Saskin}, {and} \bibinfo{person}{Michael~L Littman}.} \bibinfo{year}{2019}\natexlab{}.
\newblock \showarticletitle{Deep reinforcement learning from policy-dependent human feedback}.
\newblock \bibinfo{journal}{\emph{arXiv preprint arXiv:1902.04257}} (\bibinfo{year}{2019}).
\newblock


\bibitem[\protect\citeauthoryear{Barber and Bishop}{Barber and Bishop}{1998}]%
        {barber1998ensemble}
\bibfield{author}{\bibinfo{person}{David Barber} {and} \bibinfo{person}{Christopher~M Bishop}.} \bibinfo{year}{1998}\natexlab{}.
\newblock \showarticletitle{Ensemble learning in Bayesian neural networks}.
\newblock \bibinfo{journal}{\emph{Nato ASI Series F Computer and Systems Sciences}}  \bibinfo{volume}{168} (\bibinfo{year}{1998}), \bibinfo{pages}{215--238}.
\newblock


\bibitem[\protect\citeauthoryear{Bian, Han, Sun, Lin, Lu, He, Jiang, and Dong}{Bian et~al\mbox{.}}{2023}]%
        {bian2023chatgpt}
\bibfield{author}{\bibinfo{person}{Ning Bian}, \bibinfo{person}{Xianpei Han}, \bibinfo{person}{Le Sun}, \bibinfo{person}{Hongyu Lin}, \bibinfo{person}{Yaojie Lu}, \bibinfo{person}{Ben He}, \bibinfo{person}{Shanshan Jiang}, {and} \bibinfo{person}{Bin Dong}.} \bibinfo{year}{2023}\natexlab{}.
\newblock \showarticletitle{Chatgpt is a knowledgeable but inexperienced solver: An investigation of commonsense problem in large language models}.
\newblock \bibinfo{journal}{\emph{arXiv preprint arXiv:2303.16421}} (\bibinfo{year}{2023}).
\newblock


\bibitem[\protect\citeauthoryear{Cao, Zhao, Cheng, Shu, Liu, Liang, Zhao, and Li}{Cao et~al\mbox{.}}{2024}]%
        {cao2024survey}
\bibfield{author}{\bibinfo{person}{Yuji Cao}, \bibinfo{person}{Huan Zhao}, \bibinfo{person}{Yuheng Cheng}, \bibinfo{person}{Ting Shu}, \bibinfo{person}{Guolong Liu}, \bibinfo{person}{Gaoqi Liang}, \bibinfo{person}{Junhua Zhao}, {and} \bibinfo{person}{Yun Li}.} \bibinfo{year}{2024}\natexlab{}.
\newblock \showarticletitle{Survey on large language model-enhanced reinforcement learning: Concept, taxonomy, and methods}.
\newblock \bibinfo{journal}{\emph{arXiv preprint arXiv:2404.00282}} (\bibinfo{year}{2024}).
\newblock


\bibitem[\protect\citeauthoryear{Carta, Romac, Wolf, Lamprier, Sigaud, and Oudeyer}{Carta et~al\mbox{.}}{2023}]%
        {carta2023grounding}
\bibfield{author}{\bibinfo{person}{Thomas Carta}, \bibinfo{person}{Cl{\'e}ment Romac}, \bibinfo{person}{Thomas Wolf}, \bibinfo{person}{Sylvain Lamprier}, \bibinfo{person}{Olivier Sigaud}, {and} \bibinfo{person}{Pierre-Yves Oudeyer}.} \bibinfo{year}{2023}\natexlab{}.
\newblock \showarticletitle{Grounding large language models in interactive environments with online reinforcement learning}. In \bibinfo{booktitle}{\emph{International Conference on Machine Learning}}. PMLR, \bibinfo{pages}{3676--3713}.
\newblock


\bibitem[\protect\citeauthoryear{Cederborg, Grover, Isbell~Jr, and Thomaz}{Cederborg et~al\mbox{.}}{2015}]%
        {cederborg2015policy}
\bibfield{author}{\bibinfo{person}{Thomas Cederborg}, \bibinfo{person}{Ishaan Grover}, \bibinfo{person}{Charles~L Isbell~Jr}, {and} \bibinfo{person}{Andrea~Lockerd Thomaz}.} \bibinfo{year}{2015}\natexlab{}.
\newblock \showarticletitle{Policy Shaping with Human Teachers.}. In \bibinfo{booktitle}{\emph{IJCAI}}. \bibinfo{pages}{3366--3372}.
\newblock


\bibitem[\protect\citeauthoryear{Chen and Xu}{Chen and Xu}{2022}]%
        {chen2022policy}
\bibfield{author}{\bibinfo{person}{Jie Chen} {and} \bibinfo{person}{Wenjun Xu}.} \bibinfo{year}{2022}\natexlab{}.
\newblock \showarticletitle{Policy gradient from demonstration and curiosity}.
\newblock \bibinfo{journal}{\emph{IEEE Transactions on Cybernetics}} \bibinfo{volume}{53}, \bibinfo{number}{8} (\bibinfo{year}{2022}), \bibinfo{pages}{4923--4933}.
\newblock


\bibitem[\protect\citeauthoryear{Chu, Zhao, Weber, Li, and Wermter}{Chu et~al\mbox{.}}{2023}]%
        {chu2023accelerating}
\bibfield{author}{\bibinfo{person}{Kun Chu}, \bibinfo{person}{Xufeng Zhao}, \bibinfo{person}{Cornelius Weber}, \bibinfo{person}{Mengdi Li}, {and} \bibinfo{person}{Stefan Wermter}.} \bibinfo{year}{2023}\natexlab{}.
\newblock \showarticletitle{Accelerating reinforcement learning of robotic manipulations via feedback from large language models}.
\newblock \bibinfo{journal}{\emph{arXiv preprint arXiv:2311.02379}} (\bibinfo{year}{2023}).
\newblock


\bibitem[\protect\citeauthoryear{{Databricks}}{{Databricks}}{2023}]%
        {rag}
\bibfield{author}{\bibinfo{person}{{Databricks}}.} \bibinfo{year}{2023}\natexlab{}.
\newblock \bibinfo{title}{Retrieval-Augmented Generation (RAG)}.
\newblock
\newblock
\urldef\tempurl%
\url{https://www.databricks.com/glossary/retrieval-augmented-generation-rag}
\showURL{%
\tempurl}
\newblock
\shownote{Accessed: 2024-08-13.}


\bibitem[\protect\citeauthoryear{Devlin, Chang, Lee, and Toutanova}{Devlin et~al\mbox{.}}{2019}]%
        {Devlin2019BERTPO}
\bibfield{author}{\bibinfo{person}{Jacob Devlin}, \bibinfo{person}{Ming-Wei Chang}, \bibinfo{person}{Kenton Lee}, {and} \bibinfo{person}{Kristina Toutanova}.} \bibinfo{year}{2019}\natexlab{}.
\newblock \showarticletitle{BERT: Pre-training of Deep Bidirectional Transformers for Language Understanding}. In \bibinfo{booktitle}{\emph{North American Chapter of the Association for Computational Linguistics}}.
\newblock
\urldef\tempurl%
\url{https://api.semanticscholar.org/CorpusID:52967399}
\showURL{%
\tempurl}


\bibitem[\protect\citeauthoryear{Du, Watkins, Wang, Colas, Darrell, Abbeel, Gupta, and Andreas}{Du et~al\mbox{.}}{2023}]%
        {du2023guiding}
\bibfield{author}{\bibinfo{person}{Yuqing Du}, \bibinfo{person}{Olivia Watkins}, \bibinfo{person}{Zihan Wang}, \bibinfo{person}{C{\'e}dric Colas}, \bibinfo{person}{Trevor Darrell}, \bibinfo{person}{Pieter Abbeel}, \bibinfo{person}{Abhishek Gupta}, {and} \bibinfo{person}{Jacob Andreas}.} \bibinfo{year}{2023}\natexlab{}.
\newblock \showarticletitle{Guiding pretraining in reinforcement learning with large language models}. In \bibinfo{booktitle}{\emph{International Conference on Machine Learning}}. PMLR, \bibinfo{pages}{8657--8677}.
\newblock


\bibitem[\protect\citeauthoryear{Eschmann}{Eschmann}{2021}]%
        {eschmann2021reward}
\bibfield{author}{\bibinfo{person}{Jonas Eschmann}.} \bibinfo{year}{2021}\natexlab{}.
\newblock \showarticletitle{Reward function design in reinforcement learning}.
\newblock \bibinfo{journal}{\emph{Reinforcement Learning Algorithms: Analysis and Applications}} (\bibinfo{year}{2021}), \bibinfo{pages}{25--33}.
\newblock


\bibitem[\protect\citeauthoryear{Felicioni, Maystre, Ghiassian, and Ciosek}{Felicioni et~al\mbox{.}}{2024}]%
        {felicioni2024importance}
\bibfield{author}{\bibinfo{person}{Nicol{\`o} Felicioni}, \bibinfo{person}{Lucas Maystre}, \bibinfo{person}{Sina Ghiassian}, {and} \bibinfo{person}{Kamil Ciosek}.} \bibinfo{year}{2024}\natexlab{}.
\newblock \showarticletitle{On the Importance of Uncertainty in Decision-Making with Large Language Models}.
\newblock \bibinfo{journal}{\emph{arXiv preprint arXiv:2404.02649}} (\bibinfo{year}{2024}).
\newblock


\bibitem[\protect\citeauthoryear{Griffith, Subramanian, Scholz, Isbell, and Thomaz}{Griffith et~al\mbox{.}}{2013}]%
        {griffith2013policy}
\bibfield{author}{\bibinfo{person}{Shane Griffith}, \bibinfo{person}{Kaushik Subramanian}, \bibinfo{person}{Jonathan Scholz}, \bibinfo{person}{Charles~L Isbell}, {and} \bibinfo{person}{Andrea~L Thomaz}.} \bibinfo{year}{2013}\natexlab{}.
\newblock \showarticletitle{Policy shaping: Integrating human feedback with reinforcement learning}.
\newblock \bibinfo{journal}{\emph{Advances in neural information processing systems}}  \bibinfo{volume}{26} (\bibinfo{year}{2013}).
\newblock


\bibitem[\protect\citeauthoryear{Guo, Pleiss, Sun, and Weinberger}{Guo et~al\mbox{.}}{2017}]%
        {guo2017calibration}
\bibfield{author}{\bibinfo{person}{Chuan Guo}, \bibinfo{person}{Geoff Pleiss}, \bibinfo{person}{Yu Sun}, {and} \bibinfo{person}{Kilian~Q Weinberger}.} \bibinfo{year}{2017}\natexlab{}.
\newblock \showarticletitle{On calibration of modern neural networks}. In \bibinfo{booktitle}{\emph{International conference on machine learning}}. PMLR, \bibinfo{pages}{1321--1330}.
\newblock


\bibitem[\protect\citeauthoryear{Harrison, Ehsan, and Riedl}{Harrison et~al\mbox{.}}{2017}]%
        {harrison2017guiding}
\bibfield{author}{\bibinfo{person}{Brent Harrison}, \bibinfo{person}{Upol Ehsan}, {and} \bibinfo{person}{Mark~O Riedl}.} \bibinfo{year}{2017}\natexlab{}.
\newblock \showarticletitle{Guiding reinforcement learning exploration using natural language}.
\newblock \bibinfo{journal}{\emph{arXiv preprint arXiv:1707.08616}} (\bibinfo{year}{2017}).
\newblock


\bibitem[\protect\citeauthoryear{Hester, Vecerik, Pietquin, Lanctot, Schaul, Piot, Horgan, Quan, Sendonaris, Osband, et~al\mbox{.}}{Hester et~al\mbox{.}}{2018}]%
        {hester2018deep}
\bibfield{author}{\bibinfo{person}{Todd Hester}, \bibinfo{person}{Matej Vecerik}, \bibinfo{person}{Olivier Pietquin}, \bibinfo{person}{Marc Lanctot}, \bibinfo{person}{Tom Schaul}, \bibinfo{person}{Bilal Piot}, \bibinfo{person}{Dan Horgan}, \bibinfo{person}{John Quan}, \bibinfo{person}{Andrew Sendonaris}, \bibinfo{person}{Ian Osband}, {et~al\mbox{.}}} \bibinfo{year}{2018}\natexlab{}.
\newblock \showarticletitle{Deep q-learning from demonstrations}. In \bibinfo{booktitle}{\emph{Proceedings of the AAAI conference on artificial intelligence}}, Vol.~\bibinfo{volume}{32}.
\newblock


\bibitem[\protect\citeauthoryear{Huang, Song, Wang, Chen, and Ma}{Huang et~al\mbox{.}}{2023}]%
        {huang2023look}
\bibfield{author}{\bibinfo{person}{Yuheng Huang}, \bibinfo{person}{Jiayang Song}, \bibinfo{person}{Zhijie Wang}, \bibinfo{person}{Huaming Chen}, {and} \bibinfo{person}{Lei Ma}.} \bibinfo{year}{2023}\natexlab{}.
\newblock \showarticletitle{Look before you leap: An exploratory study of uncertainty measurement for large language models}.
\newblock \bibinfo{journal}{\emph{arXiv preprint arXiv:2307.10236}} (\bibinfo{year}{2023}).
\newblock


\bibitem[\protect\citeauthoryear{Jagodnik, Thomas, van~den Bogert, Branicky, and Kirsch}{Jagodnik et~al\mbox{.}}{2017}]%
        {jagodnik2017training}
\bibfield{author}{\bibinfo{person}{Kathleen~M Jagodnik}, \bibinfo{person}{Philip~S Thomas}, \bibinfo{person}{Antonie~J van~den Bogert}, \bibinfo{person}{Michael~S Branicky}, {and} \bibinfo{person}{Robert~F Kirsch}.} \bibinfo{year}{2017}\natexlab{}.
\newblock \showarticletitle{Training an actor-critic reinforcement learning controller for arm movement using human-generated rewards}.
\newblock \bibinfo{journal}{\emph{IEEE Transactions on Neural Systems and Rehabilitation Engineering}} \bibinfo{volume}{25}, \bibinfo{number}{10} (\bibinfo{year}{2017}), \bibinfo{pages}{1892--1905}.
\newblock


\bibitem[\protect\citeauthoryear{Janner, Li, and Levine}{Janner et~al\mbox{.}}{2021}]%
        {janner2021offline}
\bibfield{author}{\bibinfo{person}{Michael Janner}, \bibinfo{person}{Qiyang Li}, {and} \bibinfo{person}{Sergey Levine}.} \bibinfo{year}{2021}\natexlab{}.
\newblock \showarticletitle{Offline reinforcement learning as one big sequence modeling problem}.
\newblock \bibinfo{journal}{\emph{Advances in neural information processing systems}}  \bibinfo{volume}{34} (\bibinfo{year}{2021}), \bibinfo{pages}{1273--1286}.
\newblock


\bibitem[\protect\citeauthoryear{Ji, Lee, Frieske, Yu, Su, Xu, Ishii, Bang, Madotto, and Fung}{Ji et~al\mbox{.}}{2023}]%
        {ji2023survey}
\bibfield{author}{\bibinfo{person}{Ziwei Ji}, \bibinfo{person}{Nayeon Lee}, \bibinfo{person}{Rita Frieske}, \bibinfo{person}{Tiezheng Yu}, \bibinfo{person}{Dan Su}, \bibinfo{person}{Yan Xu}, \bibinfo{person}{Etsuko Ishii}, \bibinfo{person}{Ye~Jin Bang}, \bibinfo{person}{Andrea Madotto}, {and} \bibinfo{person}{Pascale Fung}.} \bibinfo{year}{2023}\natexlab{}.
\newblock \showarticletitle{Survey of hallucination in natural language generation}.
\newblock \bibinfo{journal}{\emph{Comput. Surveys}} \bibinfo{volume}{55}, \bibinfo{number}{12} (\bibinfo{year}{2023}), \bibinfo{pages}{1--38}.
\newblock


\bibitem[\protect\citeauthoryear{Jiang, Araki, Ding, and Neubig}{Jiang et~al\mbox{.}}{2021}]%
        {jiang2021can}
\bibfield{author}{\bibinfo{person}{Zhengbao Jiang}, \bibinfo{person}{Jun Araki}, \bibinfo{person}{Haibo Ding}, {and} \bibinfo{person}{Graham Neubig}.} \bibinfo{year}{2021}\natexlab{}.
\newblock \showarticletitle{How can we know when language models know? on the calibration of language models for question answering}.
\newblock \bibinfo{journal}{\emph{Transactions of the Association for Computational Linguistics}}  \bibinfo{volume}{9} (\bibinfo{year}{2021}), \bibinfo{pages}{962--977}.
\newblock


\bibitem[\protect\citeauthoryear{Kendall and Gal}{Kendall and Gal}{2017}]%
        {kendall2017uncertainties}
\bibfield{author}{\bibinfo{person}{Alex Kendall} {and} \bibinfo{person}{Yarin Gal}.} \bibinfo{year}{2017}\natexlab{}.
\newblock \showarticletitle{What uncertainties do we need in bayesian deep learning for computer vision?}
\newblock \bibinfo{journal}{\emph{Advances in neural information processing systems}}  \bibinfo{volume}{30} (\bibinfo{year}{2017}).
\newblock


\bibitem[\protect\citeauthoryear{Knox and Stone}{Knox and Stone}{2009}]%
        {knox2009interactively}
\bibfield{author}{\bibinfo{person}{W~Bradley Knox} {and} \bibinfo{person}{Peter Stone}.} \bibinfo{year}{2009}\natexlab{}.
\newblock \showarticletitle{Interactively shaping agents via human reinforcement: The TAMER framework}. In \bibinfo{booktitle}{\emph{Proceedings of the fifth international conference on Knowledge capture}}. \bibinfo{pages}{9--16}.
\newblock


\bibitem[\protect\citeauthoryear{Knox and Stone}{Knox and Stone}{2012}]%
        {knox2012reinforcement}
\bibfield{author}{\bibinfo{person}{W~Bradley Knox} {and} \bibinfo{person}{Peter Stone}.} \bibinfo{year}{2012}\natexlab{}.
\newblock \showarticletitle{Reinforcement learning from simultaneous human and MDP reward.}. In \bibinfo{booktitle}{\emph{AAMAS}}, Vol.~\bibinfo{volume}{1004}. Valencia, \bibinfo{pages}{475--482}.
\newblock


\bibitem[\protect\citeauthoryear{Lakshminarayanan, Pritzel, and Blundell}{Lakshminarayanan et~al\mbox{.}}{2017}]%
        {lakshminarayanan2017simple}
\bibfield{author}{\bibinfo{person}{Balaji Lakshminarayanan}, \bibinfo{person}{Alexander Pritzel}, {and} \bibinfo{person}{Charles Blundell}.} \bibinfo{year}{2017}\natexlab{}.
\newblock \showarticletitle{Simple and scalable predictive uncertainty estimation using deep ensembles}.
\newblock \bibinfo{journal}{\emph{Advances in neural information processing systems}}  \bibinfo{volume}{30} (\bibinfo{year}{2017}).
\newblock


\bibitem[\protect\citeauthoryear{Li, Yang, Wang, Zhu, Zhou, Qiao, Wang, Li, Lu, and Dai}{Li et~al\mbox{.}}{2024}]%
        {li2024auto}
\bibfield{author}{\bibinfo{person}{Hao Li}, \bibinfo{person}{Xue Yang}, \bibinfo{person}{Zhaokai Wang}, \bibinfo{person}{Xizhou Zhu}, \bibinfo{person}{Jie Zhou}, \bibinfo{person}{Yu Qiao}, \bibinfo{person}{Xiaogang Wang}, \bibinfo{person}{Hongsheng Li}, \bibinfo{person}{Lewei Lu}, {and} \bibinfo{person}{Jifeng Dai}.} \bibinfo{year}{2024}\natexlab{}.
\newblock \showarticletitle{Auto mc-reward: Automated dense reward design with large language models for minecraft}. In \bibinfo{booktitle}{\emph{Proceedings of the IEEE/CVF Conference on Computer Vision and Pattern Recognition}}. \bibinfo{pages}{16426--16435}.
\newblock


\bibitem[\protect\citeauthoryear{Li, Puig, Paxton, Du, Wang, Fan, Chen, Huang, Aky{\"u}rek, Anandkumar, et~al\mbox{.}}{Li et~al\mbox{.}}{2022}]%
        {li2022pre}
\bibfield{author}{\bibinfo{person}{Shuang Li}, \bibinfo{person}{Xavier Puig}, \bibinfo{person}{Chris Paxton}, \bibinfo{person}{Yilun Du}, \bibinfo{person}{Clinton Wang}, \bibinfo{person}{Linxi Fan}, \bibinfo{person}{Tao Chen}, \bibinfo{person}{De-An Huang}, \bibinfo{person}{Ekin Aky{\"u}rek}, \bibinfo{person}{Anima Anandkumar}, {et~al\mbox{.}}} \bibinfo{year}{2022}\natexlab{}.
\newblock \showarticletitle{Pre-trained language models for interactive decision-making}.
\newblock \bibinfo{journal}{\emph{Advances in Neural Information Processing Systems}}  \bibinfo{volume}{35} (\bibinfo{year}{2022}), \bibinfo{pages}{31199--31212}.
\newblock


\bibitem[\protect\citeauthoryear{Li}{Li}{2023}]%
        {li2023deep}
\bibfield{author}{\bibinfo{person}{Shengbo~Eben Li}.} \bibinfo{year}{2023}\natexlab{}.
\newblock \showarticletitle{Deep reinforcement learning}.
\newblock In \bibinfo{booktitle}{\emph{Reinforcement learning for sequential decision and optimal control}}. \bibinfo{publisher}{Springer}, \bibinfo{pages}{365--402}.
\newblock


\bibitem[\protect\citeauthoryear{Lin, Du, Watkins, Hafner, Abbeel, Klein, and Dragan}{Lin et~al\mbox{.}}{2023}]%
        {lin2023learning}
\bibfield{author}{\bibinfo{person}{Jessy Lin}, \bibinfo{person}{Yuqing Du}, \bibinfo{person}{Olivia Watkins}, \bibinfo{person}{Danijar Hafner}, \bibinfo{person}{Pieter Abbeel}, \bibinfo{person}{Dan Klein}, {and} \bibinfo{person}{Anca Dragan}.} \bibinfo{year}{2023}\natexlab{}.
\newblock \showarticletitle{Learning to model the world with language}.
\newblock \bibinfo{journal}{\emph{arXiv preprint arXiv:2308.01399}} (\bibinfo{year}{2023}).
\newblock


\bibitem[\protect\citeauthoryear{Lin, Ma, Gomez, Nakamura, He, and Li}{Lin et~al\mbox{.}}{2020}]%
        {lin2020review}
\bibfield{author}{\bibinfo{person}{Jinying Lin}, \bibinfo{person}{Zhen Ma}, \bibinfo{person}{Randy Gomez}, \bibinfo{person}{Keisuke Nakamura}, \bibinfo{person}{Bo He}, {and} \bibinfo{person}{Guangliang Li}.} \bibinfo{year}{2020}\natexlab{}.
\newblock \showarticletitle{A review on interactive reinforcement learning from human social feedback}.
\newblock \bibinfo{journal}{\emph{IEEE Access}}  \bibinfo{volume}{8} (\bibinfo{year}{2020}), \bibinfo{pages}{120757--120765}.
\newblock


\bibitem[\protect\citeauthoryear{MacGlashan, Ho, Loftin, Peng, Wang, Roberts, Taylor, and Littman}{MacGlashan et~al\mbox{.}}{2017}]%
        {macglashan2017interactive}
\bibfield{author}{\bibinfo{person}{James MacGlashan}, \bibinfo{person}{Mark~K Ho}, \bibinfo{person}{Robert Loftin}, \bibinfo{person}{Bei Peng}, \bibinfo{person}{Guan Wang}, \bibinfo{person}{David~L Roberts}, \bibinfo{person}{Matthew~E Taylor}, {and} \bibinfo{person}{Michael~L Littman}.} \bibinfo{year}{2017}\natexlab{}.
\newblock \showarticletitle{Interactive learning from policy-dependent human feedback}. In \bibinfo{booktitle}{\emph{International conference on machine learning}}. PMLR, \bibinfo{pages}{2285--2294}.
\newblock


\bibitem[\protect\citeauthoryear{Miao, Meng, Liu, Zhou, and Zhou}{Miao et~al\mbox{.}}{2021}]%
        {miao2021prevent}
\bibfield{author}{\bibinfo{person}{Mengqi Miao}, \bibinfo{person}{Fandong Meng}, \bibinfo{person}{Yijin Liu}, \bibinfo{person}{Xiao-Hua Zhou}, {and} \bibinfo{person}{Jie Zhou}.} \bibinfo{year}{2021}\natexlab{}.
\newblock \showarticletitle{Prevent the language model from being overconfident in neural machine translation}.
\newblock \bibinfo{journal}{\emph{arXiv preprint arXiv:2105.11098}} (\bibinfo{year}{2021}).
\newblock


\bibitem[\protect\citeauthoryear{Moreira, Rivas, Cruz, Dazeley, Ayala, and Fernandes}{Moreira et~al\mbox{.}}{2020}]%
        {moreira2020deep}
\bibfield{author}{\bibinfo{person}{Ithan Moreira}, \bibinfo{person}{Javier Rivas}, \bibinfo{person}{Francisco Cruz}, \bibinfo{person}{Richard Dazeley}, \bibinfo{person}{Angel Ayala}, {and} \bibinfo{person}{Bruno Fernandes}.} \bibinfo{year}{2020}\natexlab{}.
\newblock \showarticletitle{Deep reinforcement learning with interactive feedback in a human--robot environment}.
\newblock \bibinfo{journal}{\emph{Applied Sciences}} \bibinfo{volume}{10}, \bibinfo{number}{16} (\bibinfo{year}{2020}), \bibinfo{pages}{5574}.
\newblock


\bibitem[\protect\citeauthoryear{Oberdiek, Rottmann, and Gottschalk}{Oberdiek et~al\mbox{.}}{2018}]%
        {oberdiek2018classification}
\bibfield{author}{\bibinfo{person}{Philipp Oberdiek}, \bibinfo{person}{Matthias Rottmann}, {and} \bibinfo{person}{Hanno Gottschalk}.} \bibinfo{year}{2018}\natexlab{}.
\newblock \showarticletitle{Classification uncertainty of deep neural networks based on gradient information}. In \bibinfo{booktitle}{\emph{Artificial Neural Networks in Pattern Recognition: 8th IAPR TC3 Workshop, ANNPR 2018, Siena, Italy, September 19--21, 2018, Proceedings 8}}. Springer, \bibinfo{pages}{113--125}.
\newblock


\bibitem[\protect\citeauthoryear{Ouyang, Wu, Jiang, Almeida, Wainwright, Mishkin, Zhang, Agarwal, Slama, Ray, et~al\mbox{.}}{Ouyang et~al\mbox{.}}{2022}]%
        {ouyang2022training}
\bibfield{author}{\bibinfo{person}{Long Ouyang}, \bibinfo{person}{Jeffrey Wu}, \bibinfo{person}{Xu Jiang}, \bibinfo{person}{Diogo Almeida}, \bibinfo{person}{Carroll Wainwright}, \bibinfo{person}{Pamela Mishkin}, \bibinfo{person}{Chong Zhang}, \bibinfo{person}{Sandhini Agarwal}, \bibinfo{person}{Katarina Slama}, \bibinfo{person}{Alex Ray}, {et~al\mbox{.}}} \bibinfo{year}{2022}\natexlab{}.
\newblock \showarticletitle{Training language models to follow instructions with human feedback}.
\newblock \bibinfo{journal}{\emph{Advances in neural information processing systems}}  \bibinfo{volume}{35} (\bibinfo{year}{2022}), \bibinfo{pages}{27730--27744}.
\newblock


\bibitem[\protect\citeauthoryear{Savage, Wang, Gallo, Boukil, Patel, Ahmad Safavi-Naini, Soroush, and Chen}{Savage et~al\mbox{.}}{2024}]%
        {savage2024large}
\bibfield{author}{\bibinfo{person}{Thomas Savage}, \bibinfo{person}{John Wang}, \bibinfo{person}{Robert Gallo}, \bibinfo{person}{Abdessalem Boukil}, \bibinfo{person}{Vishwesh Patel}, \bibinfo{person}{Seyed~Amir Ahmad Safavi-Naini}, \bibinfo{person}{Ali Soroush}, {and} \bibinfo{person}{Jonathan~H Chen}.} \bibinfo{year}{2024}\natexlab{}.
\newblock \showarticletitle{Large Language Model Uncertainty Measurement and Calibration for Medical Diagnosis and Treatment}.
\newblock \bibinfo{journal}{\emph{medRxiv}} (\bibinfo{year}{2024}), \bibinfo{pages}{2024--06}.
\newblock


\bibitem[\protect\citeauthoryear{Schulman, Wolski, Dhariwal, Radford, and Klimov}{Schulman et~al\mbox{.}}{2017}]%
        {schulman2017proximal}
\bibfield{author}{\bibinfo{person}{John Schulman}, \bibinfo{person}{Filip Wolski}, \bibinfo{person}{Prafulla Dhariwal}, \bibinfo{person}{Alec Radford}, {and} \bibinfo{person}{Oleg Klimov}.} \bibinfo{year}{2017}\natexlab{}.
\newblock \showarticletitle{Proximal policy optimization algorithms}.
\newblock \bibinfo{journal}{\emph{arXiv preprint arXiv:1707.06347}} (\bibinfo{year}{2017}).
\newblock


\bibitem[\protect\citeauthoryear{Science and on~Artificial~Intelligence}{Science and on~Artificial~Intelligence}{2019}]%
        {national2019national}
\bibfield{author}{\bibinfo{person}{National Science} {and} \bibinfo{person}{Technology Council (US). Select~Committee on Artificial~Intelligence}.} \bibinfo{year}{2019}\natexlab{}.
\newblock \bibinfo{booktitle}{\emph{The National Artificial Intelligence Research and Development Strategic Plan: 2023 Update}}.
\newblock \bibinfo{publisher}{National Science and Technology Council (US), Select Committee on Artificial~…}.
\newblock


\bibitem[\protect\citeauthoryear{Shi, Liu, Ze, Du, and Xu}{Shi et~al\mbox{.}}{2023}]%
        {shi2023unleashing}
\bibfield{author}{\bibinfo{person}{Ruizhe Shi}, \bibinfo{person}{Yuyao Liu}, \bibinfo{person}{Yanjie Ze}, \bibinfo{person}{Simon~S Du}, {and} \bibinfo{person}{Huazhe Xu}.} \bibinfo{year}{2023}\natexlab{}.
\newblock \showarticletitle{Unleashing the power of pre-trained language models for offline reinforcement learning}.
\newblock \bibinfo{journal}{\emph{arXiv preprint arXiv:2310.20587}} (\bibinfo{year}{2023}).
\newblock


\bibitem[\protect\citeauthoryear{Shou and Di}{Shou and Di}{2020}]%
        {shou2020reward}
\bibfield{author}{\bibinfo{person}{Zhenyu Shou} {and} \bibinfo{person}{Xuan Di}.} \bibinfo{year}{2020}\natexlab{}.
\newblock \showarticletitle{Reward design for driver repositioning using multi-agent reinforcement learning}.
\newblock \bibinfo{journal}{\emph{Transportation research part C: emerging technologies}}  \bibinfo{volume}{119} (\bibinfo{year}{2020}), \bibinfo{pages}{102738}.
\newblock


\bibitem[\protect\citeauthoryear{Singh, Blukis, Mousavian, Goyal, Xu, Tremblay, Fox, Thomason, and Garg}{Singh et~al\mbox{.}}{2023}]%
        {singh2023progprompt}
\bibfield{author}{\bibinfo{person}{Ishika Singh}, \bibinfo{person}{Valts Blukis}, \bibinfo{person}{Arsalan Mousavian}, \bibinfo{person}{Ankit Goyal}, \bibinfo{person}{Danfei Xu}, \bibinfo{person}{Jonathan Tremblay}, \bibinfo{person}{Dieter Fox}, \bibinfo{person}{Jesse Thomason}, {and} \bibinfo{person}{Animesh Garg}.} \bibinfo{year}{2023}\natexlab{}.
\newblock \showarticletitle{ProgPrompt: program generation for situated robot task planning using large language models}.
\newblock \bibinfo{journal}{\emph{Autonomous Robots}} \bibinfo{volume}{47}, \bibinfo{number}{8} (\bibinfo{year}{2023}), \bibinfo{pages}{999--1012}.
\newblock


\bibitem[\protect\citeauthoryear{Suay, Brys, Taylor, and Chernova}{Suay et~al\mbox{.}}{2016}]%
        {suay2016learning}
\bibfield{author}{\bibinfo{person}{Halit~Bener Suay}, \bibinfo{person}{Tim Brys}, \bibinfo{person}{Matthew~E Taylor}, {and} \bibinfo{person}{Sonia Chernova}.} \bibinfo{year}{2016}\natexlab{}.
\newblock \showarticletitle{Learning from demonstration for shaping through inverse reinforcement learning}. In \bibinfo{booktitle}{\emph{Proceedings of the 2016 international conference on autonomous agents \& multiagent systems}}. \bibinfo{pages}{429--437}.
\newblock


\bibitem[\protect\citeauthoryear{Tamkin, Brundage, Clark, and Ganguli}{Tamkin et~al\mbox{.}}{2021}]%
        {tamkin2021understanding}
\bibfield{author}{\bibinfo{person}{Alex Tamkin}, \bibinfo{person}{Miles Brundage}, \bibinfo{person}{Jack Clark}, {and} \bibinfo{person}{Deep Ganguli}.} \bibinfo{year}{2021}\natexlab{}.
\newblock \showarticletitle{Understanding the capabilities, limitations, and societal impact of large language models}.
\newblock \bibinfo{journal}{\emph{arXiv preprint arXiv:2102.02503}} (\bibinfo{year}{2021}).
\newblock


\bibitem[\protect\citeauthoryear{Tasrin, Nahian, Perera, and Harrison}{Tasrin et~al\mbox{.}}{2021}]%
        {tasrin2021influencing}
\bibfield{author}{\bibinfo{person}{Tasmia Tasrin}, \bibinfo{person}{Md~Sultan~Al Nahian}, \bibinfo{person}{Habarakadage Perera}, {and} \bibinfo{person}{Brent Harrison}.} \bibinfo{year}{2021}\natexlab{}.
\newblock \showarticletitle{Influencing reinforcement learning through natural language guidance}.
\newblock \bibinfo{journal}{\emph{arXiv preprint arXiv:2104.01506}} (\bibinfo{year}{2021}).
\newblock


\bibitem[\protect\citeauthoryear{Thirunavukarasu, Ting, Elangovan, Gutierrez, Tan, and Ting}{Thirunavukarasu et~al\mbox{.}}{2023}]%
        {thirunavukarasu2023large}
\bibfield{author}{\bibinfo{person}{Arun~James Thirunavukarasu}, \bibinfo{person}{Darren Shu~Jeng Ting}, \bibinfo{person}{Kabilan Elangovan}, \bibinfo{person}{Laura Gutierrez}, \bibinfo{person}{Ting~Fang Tan}, {and} \bibinfo{person}{Daniel Shu~Wei Ting}.} \bibinfo{year}{2023}\natexlab{}.
\newblock \showarticletitle{Large language models in medicine}.
\newblock \bibinfo{journal}{\emph{Nature medicine}} \bibinfo{volume}{29}, \bibinfo{number}{8} (\bibinfo{year}{2023}), \bibinfo{pages}{1930--1940}.
\newblock


\bibitem[\protect\citeauthoryear{Warnell, Waytowich, Lawhern, and Stone}{Warnell et~al\mbox{.}}{2018}]%
        {warnell2018deep}
\bibfield{author}{\bibinfo{person}{Garrett Warnell}, \bibinfo{person}{Nicholas Waytowich}, \bibinfo{person}{Vernon Lawhern}, {and} \bibinfo{person}{Peter Stone}.} \bibinfo{year}{2018}\natexlab{}.
\newblock \showarticletitle{Deep tamer: Interactive agent shaping in high-dimensional state spaces}. In \bibinfo{booktitle}{\emph{Proceedings of the AAAI conference on artificial intelligence}}, Vol.~\bibinfo{volume}{32}.
\newblock


\bibitem[\protect\citeauthoryear{Wei, Wei, Tay, Tran, Webson, Lu, Chen, Liu, Huang, Zhou, et~al\mbox{.}}{Wei et~al\mbox{.}}{2023}]%
        {wei2023larger}
\bibfield{author}{\bibinfo{person}{Jerry Wei}, \bibinfo{person}{Jason Wei}, \bibinfo{person}{Yi Tay}, \bibinfo{person}{Dustin Tran}, \bibinfo{person}{Albert Webson}, \bibinfo{person}{Yifeng Lu}, \bibinfo{person}{Xinyun Chen}, \bibinfo{person}{Hanxiao Liu}, \bibinfo{person}{Da Huang}, \bibinfo{person}{Denny Zhou}, {et~al\mbox{.}}} \bibinfo{year}{2023}\natexlab{}.
\newblock \showarticletitle{Larger language models do in-context learning differently}.
\newblock \bibinfo{journal}{\emph{arXiv preprint arXiv:2303.03846}} (\bibinfo{year}{2023}).
\newblock


\bibitem[\protect\citeauthoryear{Wen, Lin, Wang, Yang, Wen, Mai, Wang, Zhang, and Zhang}{Wen et~al\mbox{.}}{2023}]%
        {wen2023large}
\bibfield{author}{\bibinfo{person}{Muning Wen}, \bibinfo{person}{Runji Lin}, \bibinfo{person}{Hanjing Wang}, \bibinfo{person}{Yaodong Yang}, \bibinfo{person}{Ying Wen}, \bibinfo{person}{Luo Mai}, \bibinfo{person}{Jun Wang}, \bibinfo{person}{Haifeng Zhang}, {and} \bibinfo{person}{Weinan Zhang}.} \bibinfo{year}{2023}\natexlab{}.
\newblock \showarticletitle{Large sequence models for sequential decision-making: a survey}.
\newblock \bibinfo{journal}{\emph{Frontiers of Computer Science}} \bibinfo{volume}{17}, \bibinfo{number}{6} (\bibinfo{year}{2023}), \bibinfo{pages}{176349}.
\newblock


\bibitem[\protect\citeauthoryear{Xiao and Wang}{Xiao and Wang}{2021}]%
        {xiao2021hallucination}
\bibfield{author}{\bibinfo{person}{Yijun Xiao} {and} \bibinfo{person}{William~Yang Wang}.} \bibinfo{year}{2021}\natexlab{}.
\newblock \showarticletitle{On hallucination and predictive uncertainty in conditional language generation}.
\newblock \bibinfo{journal}{\emph{arXiv preprint arXiv:2103.15025}} (\bibinfo{year}{2021}).
\newblock


\bibitem[\protect\citeauthoryear{Yao, Rao, Hausknecht, and Narasimhan}{Yao et~al\mbox{.}}{2020}]%
        {yao2020keep}
\bibfield{author}{\bibinfo{person}{Shunyu Yao}, \bibinfo{person}{Rohan Rao}, \bibinfo{person}{Matthew Hausknecht}, {and} \bibinfo{person}{Karthik Narasimhan}.} \bibinfo{year}{2020}\natexlab{}.
\newblock \showarticletitle{Keep calm and explore: Language models for action generation in text-based games}.
\newblock \bibinfo{journal}{\emph{arXiv preprint arXiv:2010.02903}} (\bibinfo{year}{2020}).
\newblock


\bibitem[\protect\citeauthoryear{Yu, Yang, Zhu, Li, et~al\mbox{.}}{Yu et~al\mbox{.}}{2018}]%
        {yu2018learning}
\bibfield{author}{\bibinfo{person}{Chao Yu}, \bibinfo{person}{Tianpei Yang}, \bibinfo{person}{Wenxuan Zhu}, \bibinfo{person}{Guangliang Li}, {et~al\mbox{.}}} \bibinfo{year}{2018}\natexlab{}.
\newblock \showarticletitle{Learning shaping strategies in human-in-the-loop interactive reinforcement learning}.
\newblock \bibinfo{journal}{\emph{arXiv preprint arXiv:1811.04272}} (\bibinfo{year}{2018}).
\newblock


\bibitem[\protect\citeauthoryear{Yu}{Yu}{2018}]%
        {yu2018towards}
\bibfield{author}{\bibinfo{person}{Yang Yu}.} \bibinfo{year}{2018}\natexlab{}.
\newblock \showarticletitle{Towards Sample Efficient Reinforcement Learning.}. In \bibinfo{booktitle}{\emph{IJCAI}}. \bibinfo{pages}{5739--5743}.
\newblock


\end{thebibliography}


\end{document}